\documentclass{llncs}
\pdfoutput=1
\usepackage{url}
\usepackage{latexsym}
\usepackage{xspace}
\usepackage{graphicx}
\usepackage{cite}
\usepackage{multirow}
\usepackage{comment}
\usepackage{soul}
\usepackage{booktabs}
\usepackage{array}

\newcommand{\ie}{\emph{i.e.,}\xspace}

\newcommand{\etal}{\emph{et al.}\xspace}

\newcommand{\nome}{\textsc{SUAEx}\xspace}

\begin{document}
\title{Simple Unsupervised Similarity-Based \\Aspect Extraction}

\author{Danny Suarez Vargas \and Lucas R. C. Pessutto \and
Viviane Pereira Moreira}

\authorrunning{D. S. Vargas \etal}
\institute{Federal University of Rio Grande do Sul -- UFRGS, Porto Alegre RS, Brazil\\
\url{http://www.inf.ufrgs.br/}\\
\email{\{dsvargas,lrcpessutto,viviane\}@inf.ufrgs.br}}

\maketitle
\begin{abstract}
In the context of sentiment analysis, there has been growing interest in performing a finer granularity analysis focusing on the specific aspects of the entities being evaluated.
This is the goal of Aspect-Based Sentiment Analysis (ABSA) which basically involves two tasks: aspect extraction and polarity detection.
The first task is responsible for discovering the aspects mentioned in the review text and the second task assigns a sentiment orientation (positive, negative, or neutral) to that aspect.
Currently, the state-of-the-art in ABSA consists of the application of deep learning methods such as recurrent, convolutional and attention neural networks.
The limitation of these techniques is that they require a lot of training data and are computationally expensive.
In this paper, we propose a simple approach called \nome for aspect extraction.
\nome is unsupervised and relies solely on the similarity of word embeddings. 
Experimental results on datasets from three different domains have shown that \nome achieves results that can outperform the state-of-the-art attention-based approach at a fraction of the time. 
  
\end{abstract}

\section{Introduction}
\vspace{-0.3cm}
Opinionated texts are abundant on the Web and its study has drawn a lot of attention from both companies and academics originating the research field known as opinion mining or sentiment analysis.
The last few years have been very prolific in this field which combines Natural Language Processing and Data Mining.
In a study which dates back to 2013, Feldman~\cite{Feldman2013} mentioned that over 7,000 articles had already been written about the topic.
Several facets of the problem have been explored and numerous solutions have been proposed.

While most of the work in the area has been devoted to assigning a polarity score (positive, negative, or neutral) to the overall sentiment conveyed by the text of an entire review, in the last few years, there has been increasing interest in performing a finer-grained analysis. Such analysis, known as \emph{Aspect-Based Sentiment Analysis} (ABSA)~\cite{zhang2014} deals basically with the texts of extracting and scoring the opinion expressed towards an entity. For example, in the sentence ``\emph{The decor of the restaurant is amazing and the food was incredible}", the words \emph{decor} and \emph{food} are the aspects of the entity (or category) \emph{restaurant}.

ABSA is a challenging task because it needs to accurately extract and rate fine-grained information from reviews. Review texts can be ambiguous and contain acronyms, slangs and misspellings.
Furthermore, aspects vary from one domain to another -- one word that represents a valid aspect in one domain, may not do so for another domain. For example, consider the input sentence \emph{``The decor of the place is really beautiful, my sister loved it."} in the \textit{Restaurant} domain. The ABSA solution should focus its attention on the words \emph{``decor"} and \emph{``restaurant"}. However, if we do not explicitly set \textit{Restaurant} as the domain, the word \emph{``sister"} can also gain attention as an aspect term. 
This poses a potential problem to extraction approaches. For example, an approach that is only based on rules~\cite{Qiu2011} can assume that aspects are always nouns and come exactly after the adjective. However, rigid rules can create false positives because not always a noun represents an aspect. Another approach could consider the distribution of words in texts and hypothesize that the number of occurrences of a given word is determinant to decide whether it is an aspect. Again, this could lead to false positives since high-frequency words tend to be stopwords.

A disadvantage of current state-of-the-art approaches~\cite{Yin2016,wang2016,wang2017coupled,Giannakopoulos2017}  is that they rely on techniques that require significant computational power, such as deep neural networks. 
In special, neural attention mechanisms~\cite{BahdanauCB14} are typically expensive.

In this paper, we propose \nome, a simple unsupervised method for aspect extraction. \nome relies on vector similarity to emulate the attention mechanism which that allows us to focus on the relevant information. 
Our main contribution is to show that a simple and inexpensive solution can perform as well as the neural attention mechanisms.
We tested \nome on datasets from different domains, and it was able to outperform the state-of-the-art in ABSA in many cases in terms of quality and in all cases in terms of time.
%

\section{Background and Definitions} \label{sec:background}
\vspace{-0.3cm}

\textbf{Word-Embeddings}. 
Representing words in a vector space is widely used as a means to map the semantic similarity between them. The underlying concept is the hypothesis that words with similar meanings are used in similar  contexts. Word embeddings are a low dimensional vector representation of words which is able to keep the distributional similarity between them. Furthermore, word embeddings are able to map some linguistic regularities present in documents. Since the original proposal by Mikolov~\etal~\cite{mikolov2013}, other techniques have been presented \cite{PenningtonSM14,BojanowskiGJM16} adding to the popularity of word embeddings.

\noindent\textbf{Category and Aspect}.
The terms \emph{category} and \emph{aspect} are defined as follows. For a given sentence $S={w_1, w_2, ...,w_n}$ taken from the text of a review, the category $C$ is the broad, general topic of $S$, while the aspects are the attributes or characteristics of $C$ \cite{liu2012sentiment}. In other words, a  category (\ie laptop, restaurant, cell phone) can be treated as a cluster of related aspects (\ie memory, battery, and processor are aspects that characterize the category \emph{laptop}).

\noindent\textbf{Reference Words} are important in the context of our proposal because they aid in the correct extraction of aspects (\ie distinguishing aspects from non-aspects), and help determine whether an aspect belongs to a category. A reference word can be an aspect, or the name of the category itself (a synonym, meronym, or hyponym).     For example, if we want to discover aspects from the category ``laptop", the words ``computer", ``pc" and the word ``laptop" itself would be reference words.

\noindent\textbf{Attention Mechanism}. The attention mechanism was introduced in a neural machine translation solution~\cite{BahdanauCB14}. The main idea was to modify the encoder-decoder structure in order to improve the performance for long sentences. For a given input sentence $S=\{w_1, w_2, ..,w_n\}$, the encoded value $e_s$  of $S$ and a set of hidden layers $H=\{h_1, h_2, ...,h_m\}$, the decoder for each output $y_i$  considers not only the value of the previous hidden layer $h_{i-1}$ and a general context $c$, but it also considers the relative importance of each output word $y_i$ with respect to the input sentence $S$. For example, for a given output word $y_j$, it can be more important to see the words $w_2, w_3$ in $S$, while for another output word $y_k$, it can be more important to see only the word $w_4$ in $S$. The attribution of the relative importance is performed by an attention mechanism.

\section{Related Work}\label{sec:related}
\vspace{-0.3cm}

Sentiment Analysis can be performed at different levels of granularity. 
One could be interested in a coarse-grained analysis which assigns a sentiment polarity to an entire review document (\ie document level analysis); or to each sentence of the review; or, at a finer granularity, to the individual aspects of an entity. 

The aspect level is quickly gaining importance, mainly due to the relevant information that it conveys~\cite{liu2012sentiment}.
In this level of analysis, the aspects and entities are identified in natural language texts. 
Aspect extraction task can be classified into three main groups according to the underlying approach ~\cite{zhang2014}:
($i$) based on language rules~\cite{hu2004mining,Qiu2011,poria2014}, 
($ii$) based on sequence labeling models~\cite{jin2009novel,jakob2010extracting}, and 
($iii$) based on topic models~\cite{moghaddam2011ilda}. 
However, other works do not fit in only one of these groups as they combine resources from more than one approach~\cite{toh2016nlangp}. 
Furthermore, state-of-the-art approaches rely on more sophisticated architectures like recurrent neural networks such as LSTM, Bi-LSTM, Neural Attention Models, and Convolutional Neural Networks~\cite{wang2016,wang2017coupled,Giannakopoulos2017,he2017, PORIA2016}.

The work proposed by He \etal~\cite{he2017}, known as Attention-based Aspect Extraction (ABAE), represents the state-of-the-art in ABSA and was used as the baseline of our work. 
ABAE relies on an attention neural network to highlight the most important words in a given text by de-emphasizing the irrelevant words. 
ABAE is a three-layer neural network. The input layer receives a given sentence $S = \{w_1, w_2, ..., w_n\}$, $S$ is represented as a set of fixed length vectors $e = \{e_w1, e_w2, ..., e_wn\}$. These vectors are processed by the hidden layer setting the attention values $a = \{a_1, a_2, ..., a_n\}$ related to a given context $y_s$. The context $y_s$ is obtained from the average of the word vectors in $e$. After performing the attention mechanism, the input sentence is encoded as $z_s$ and a dimensionality reduction is performed from the word-embedding space to the aspect-embedding space. In other words, the input sentence is represented only by the most relevant words in $r_s$. In addition, the process of training a neural network needs an optimization function. The output layer is the sentence reconstruction $r_s$ and the function to optimize aims to maximize the similarity between $z_s$ and $r_s$. 
Finally, the mathematical definition of ABAE is the following:

\begin{small}
\centering
        $r_s$ = $T^T$.$p_t$ ~~~~
         $p_t$ = softmax($W$.$z_s$+ $b$)\\
            $z_s$ = $\sum_{i=1}^{n}a_i e_{w_i}$ ~~~~
        $a_i$ = $\frac{exp(d_i)}{\sum_{j=1}^{n}exp(d_j)}$\\
        $d_i$ = $e_{w_i}^{T}My_s$ ~~~~
        $y_s$ = $\frac{1}{n}\sum_{i=1}^{n}e_{w_i}$\\
        $J(\theta) = \sum_{s \in D}\sum_{i=1}^{m}max(0, 1 - r_sz_s + r_sn_i)$\\
\end{small}
\noindent where $z_s$ encodes the input sentence $S$ by considering the relevance of its words. $a_i$ is the relevance of the $i^{th}$ word in $S$. $d_i$ is the value that expresses the importance of the $i^{th}$  word related to the context $y_s$. Finally, $J(\theta)$ is the objective function which is optimized in the training process. 

%
In summary, each group of solutions for ABSA have advantages and disadvantages. The methods based on language rules are simple but require manual annotation to construct the initial set of rules. Furthermore, these rules are domain-specific -- a new set of rules is needed for each domain. The methods based on sequence models, topic models and even some based on neural networks are supervised machine learning solutions. So, their quality is directly proportional to the amount of annotated data. Finally, methods based on unsupervised neural networks, such as our baseline~\cite{he2017}, achieve good results but at a high computational cost. 

\vspace{-0.3cm}
\section{Simple Unsupervised Aspect Extraction}\label{sec:suaex}
\vspace{-0.3cm}

This section introduces \nome, a simple unsupervised similarity-based solution for ABSA. \nome relies on the similarity of vector representations of words to emulate the attention mechanism used in the state-of-the-art. Since our proposed solution does not need to train a neural network, \nome is computationally cheaper than state-of-the-art solutions and, as demonstrated in Section~\ref{sec:results}, it achieves results that can surpass the baseline at a fraction of the time. 


\begin{figure}[ht]
    \begin{center}
          \includegraphics[width=1.0\textwidth]{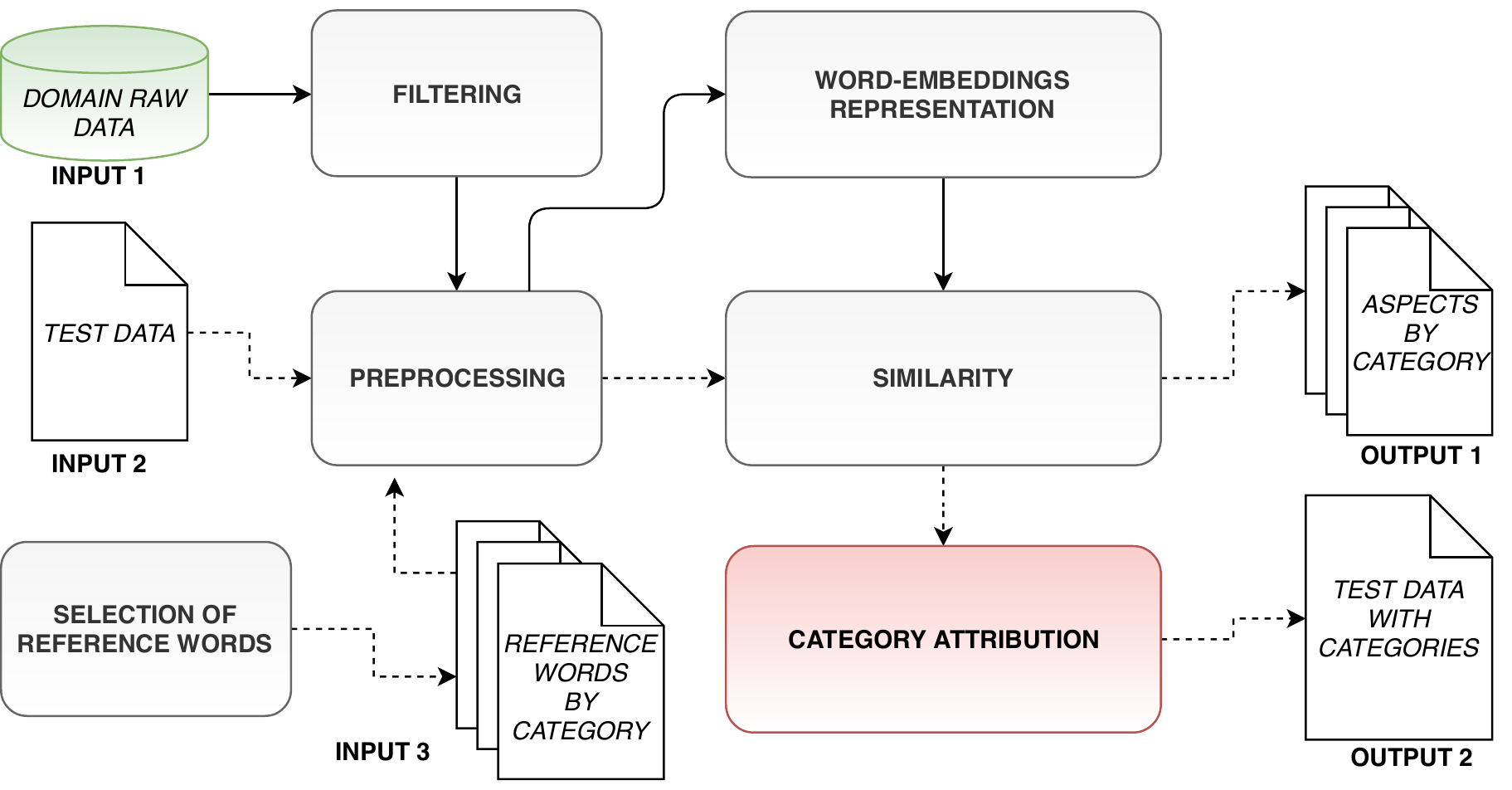}
            \caption{\nome framework\label{d_suaex}. The continuous arrows represent the path taken by Input 1, while the dashed arrows represent the path followed by Inputs 2 and 3.}
    \end{center}
    \vspace{-0.4cm}
\end{figure}

\nome\footnote{The code for \nome is available at \url{https://github.com/dannysvof/SUAEx.git}.} consists in six modules depicted in Fig.~\ref{d_suaex}: ($i$) Filtering, ($ii$) Selection of Reference Words, ($iii$) Preprocessing, ($iv$) Word-Embeddings Representation, ($v$) Similarity, and ($vi$) Category Attribution. 
\nome requires three inputs and generates two outputs. 
The inputs are: 
\textbf{Input1} -- the \emph{raw data} expressed as free text from a given domain (which is used to build the domain-specific word embeddings);
\textbf{Input2} -- the \emph{test data} with the reviews for which the aspects and categories will be extracted; and
\textbf{Input3} -- the \emph{reference words}, which are used to determine the categories as well as extract the aspects related to each category.
Next, we describe the components of \nome.

The \textbf{Selection of Reference Words} module is responsible for choosing the representative words for each category.  
In other words, if we want $k$ categories as output, we need to select $k$ groups of reference words. The selection of the words for each group can be performed in three different ways: manual, semi-automatic, and automatic. The manual selection can be done by simply selecting the category words themselves as reference words. The semi-automatic selection can be performed by expanding the initial manually constructed groups of reference words. The expansion can be done through the search for synonyms or meronyms of the words that represent the category name. Finally, an automatic selection mechanism can be performed by considering a taxonomy of objects~\cite{Salazar2014}. 

The \textbf{Filtering} module aims to select the domain related part from raw data. This module is optional but it is particularly useful when we want to delve into a certain category and we only have raw data for the general topic. For example, if we just performed the aspect extraction for the category ``Electronics", we have the raw data for it. If now, we want to perform the aspect extraction of the category ``Laptop", we need only raw data about this new more specific category. This module is in charge of selecting the right raw data from a large corpus, thus manual filtering is unfeasible. Filtering can be implemented as a binary text classifier for the domain of interest, or simply by choosing reviews that mention the category name.

The \textbf{Preprocessing} module normalizes the input data and reduces the amount of raw data needed to construct the vector representation of words (word-embeddings). The amount of data needed to train word-embeddings is directly proportional to the size of the vocabulary of the raw data. Since preprocessing reduces the size of the vocabulary, it has the effect of reducing the amount of raw data needed.  This module encompasses typical preprocessing tasks such as tokenization, sentence splitting, lemmatization, and stemming. 

The \textbf{Word-Embedding Representation} module is responsible for creating the vector representation of words which will be used to measure word similarity in the Similarity module. It receives the preprocessed raw data, transforms it into a vector representation and returns a domain-specific model. The model can be generated using well-known tools such as Word2vec~\cite{mikolov2013}, Glove~\cite{PenningtonSM14}, or Fastext~\cite{BojanowskiGJM16}. This module is particularly important because \nome relies solely on the similarity of domain-specific word-embeddings. 

The \textbf{Similarity as an Attention Mechanism} module receives two types of inputs, the preprocessed reference words and the test data. The goal of this module is to emulate the behavior of the attention mechanism in a neural network by assigning attention values to each word in an input sentence in relation to a given set of reference words. For each group of reference words, (which are in the same number as the categories desired as output), it returns an attention-valued version of the test data. This output can be used in two ways: to identify the aspects for each category or as an input to the Category Attribution module. A vector similarity measure, like the cosine similarity, is used to attribute the relevance of a given word $x$ in relation to another word $y$ or related to a group of words $c$. In this module, we can test with two types of similarity values. The similarity obtained from the direct comparison of two words (direct similarity) and the similarity obtained from the comparison of two words in relation to some contextual words (contextual similarity). Finally, the attention values are obtained by applying the $softmax$ function to the similarity values.

\textbf{Output1} is the test data together with the values for attention and similarity assigned by \nome. For example, if we consider three groups of reference words in the input, "food", "staff", and "ambience", Output1 consists in three attention-valued and similarity-valued versions of the test data (one set of values for each category).

\textbf{The Category Attribution} module uses the output of the Similarity module to assign one of the desired categories to each sentence in the test data. In this module, we can test different ways to aggregate the similarity values assigned to the words. 
For example, one could use the average for each sentence or only consider the maximum value~\cite{VargasM17}. If the average is used, it means that there are more than one relevant word to receive attention in the sentence. However, if only the maximum value is used the word with the highest score will get all the attention.

\textbf{Output2} is the main output which contains all the sentences of the test data with the categories assigned by the Category Attribution module.

\vspace{-0.3cm}
\section{Experimental Design}\label{sec:experiments}
\vspace{-0.3cm}
The experiments of this aim at evaluating \nome for aspect and category extraction both in terms of quality and runtime. Our tests are done over datasets coming from three different domains. 

The experiments are organized in two parts. In the first experiment, the goal is to compare \nome to our baseline ABAE~\cite{he2017}.We hope to answer the following questions:
\emph{Can a simple approach like \nome achieve results that are close to the state-of-the-art in aspect extraction?} and \emph{How does \nome behave in different domains?}
The second experiment performs a runtime analysis of the two approaches.
Below, we describe the datasets, the tools and resources used. 

\vspace{0.2cm}
\noindent\textbf{Datasets}.
The datasets used in our experiments are summarized in Table~\ref{tab:datasets} 
and come from different sources and different domains to enable a broad evaluation. They are all freely available to allow comparison with existing approaches.\\
\begin{table}[h!]
    \centering
    \caption{Dataset Statistics}
    \begin{tabular}{l|l|r}
    \hline
    \bf{Dataset Name} & \bf{file} & \bf{\# sentences}\\
    \hline 
    CitySearch  & train    & 281,989  \\
    CitySearch  & test     & 3,328 \\
    BeerAdvocate        & train    & 16,882,700\\
    BeerAdvocate       & test      & 9,236\\
    \hline 
    Sem2015-Restaurant   & train & 281,989 \\
    Sem2015-Restaurant   & test  & 453 \\
    Sem2015-Laptop      & train & 1,083,831 \\
    Sem2015-Laptop      & test  & 241 \\
    \hline 
    \end{tabular}
    \label{tab:datasets}
\vspace{-0.3cm}
\end{table}
\textit{ABAE datasets}
One of our sources of data is our baseline~\cite{he2017}, which made two datasets available -- one in the domain of \emph{Restaurant}, known as \textit{CitySearch} and another in the domain of \emph{Beer} (originally presented in~\cite{McAuley2012}), known as \emph{BeerAdvocate}. Each dataset consists of two files: one for training the vector representation of words and one with test data. 
We used these datasets to test and compare our method to the baseline under the same conditions.
For each sentence in the test file, the datasets have annotations that indicate the expected category.\\
\textit{SemEval datasets} 
Our second source of data is the SemEval evaluation campaigns, specifically from an ABSA task\footnote{\url{http://alt.qcri.org/semeval2015/task12/}}. 
The reviews are on the domains of \emph{Restaurant} and 
\emph{Laptop}. We used the train and test files which contain the text of the reviews, the Aspect Category, as well as the aspect words, aspect word position in text, and their polarity. In our experiments, we considered the category entities as the categories for each review text. For example, if the review text is \emph{``The pizza was great"}, the category for the aspect cluster is the word \emph{``food"}. 
The SemEval datasets 
were used with the goal of testing adaptability of our solution across domains. 
In order to run the \nome with the SemEval datasets, some modifications on the \emph{test data} had to be made: 
($i$) we removed the instances which could be classified more than one category;
($ii$) we considered only the entities as category labels; and
($iii$) we discarded the categories with very few instances.
The same modified version was submitted to both competing systems.
For the datasets 
on the \emph{Restaurant} domain, we selected three groups of reference words, $\{``food"\}$, $\{``staff"\}$, and $\{``ambience"\}$. 
For BeerAdvocate, we selected the groups $\{``feel"\}$, $\{``taste"\}$, $\{``look"\}$, and $\{``overall"\}$. And for the dataset Sem2015-Laptop, we selected the groups $\{``price"\}$, $\{``hardware"\}$, $\{``software"\}$, and $\{``support"\}$. 

\vspace{0.2cm}
\noindent\textbf{Tools and Parametrization}.
NLTK\footnote{\url{https://www.nltk.org/}} was used in the preprocessing module to remove stopwords, perform tokenization, sentence splitting, and for lemmatization. 
The domain-specific word-embeddings were created with Word2Vec\footnote{\url{https://code.google.com/archive/p/word2vec/}} using the following configuration: CBOW module, window size of 5 words, and 200 dimensions for the resulting vectors. The remaining parameters were used with the default values (negative sampling = 5, number of iterations = 15).
The similarity between word vectors was measured with the cosine similarity in Gensim~\cite{gensim_rehurek_lrec} which reads the model created by the Word2vec. 
Scikit-learn\footnote{\url{https://scikit-learn.org/stable/}} provided the metrics for evaluating the quality of the aspects and categories extracted with the traditional  metrics (precision, recall, and F1).  
\emph{Amazon raw data} was taken from a public repository\footnote{\url{http://jmcauley.ucsd.edu/data/amazon/}} to be used as an external source of data. This data was necessary because for the \emph{laptop} domain, since the training file provided by SemEval is too small and insufficient to create the domain word-embeddings.

\vspace{0.2cm}
\noindent\textbf{Baseline}.
Our baseline, ABAE~\cite{he2017}, was summarized in Section~\ref{sec:related}. Its code is available on a github repository\footnote{\url{https://github.com/ruidan/Unsupervised-Aspect-Extraction}}. We downloaded it and ran the experiments using the default configurations  according to the authors' instructions (\ie  word-embeddings dimension = 200, batch-size = 50, vocabulary size = 9000, aspect groups = 14, training epochs = 15). Despite the authors having released a pre-trained model for the restaurant domain, we ran the provided code from scratch and step-by-step in order to measure the execution time.
%
\vspace{-0.2cm}
\section{Results and Discussion}\label{sec:results}
\begin{figure}[h!]
    \begin{center}
          \includegraphics[width=\textwidth]{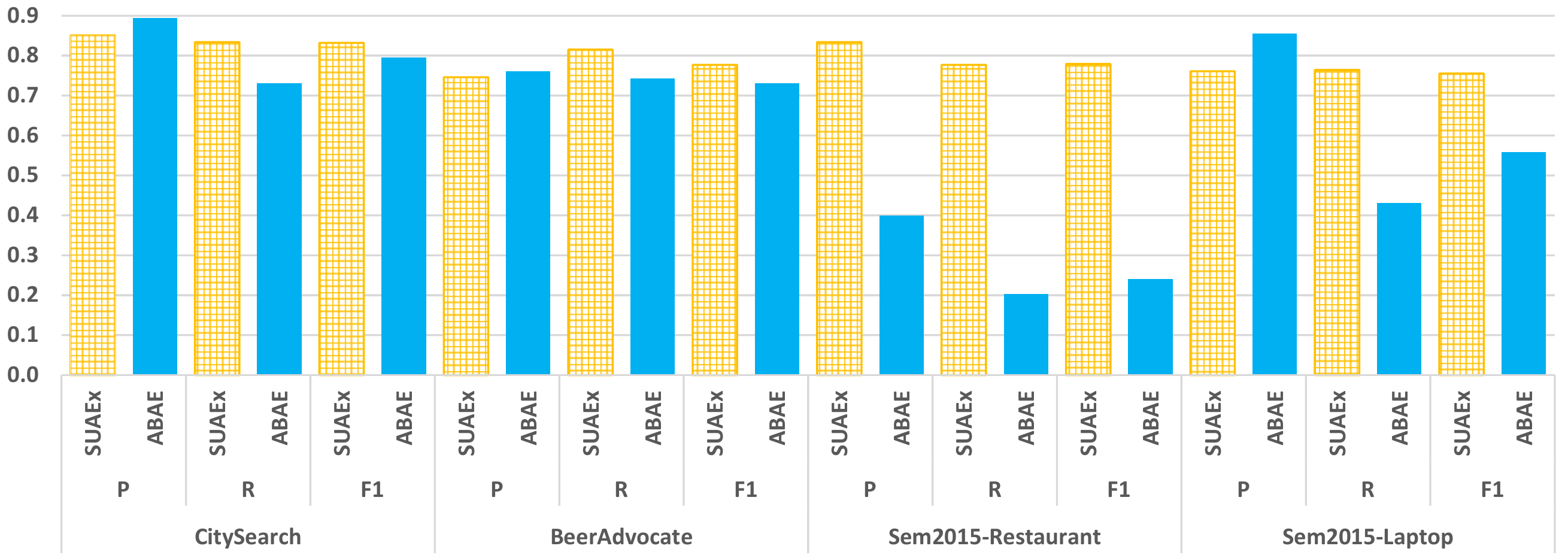}
            \caption{Overall results averaged across categories \label{im:overall}}
            \vspace{-0.3cm}
    \end{center}
\end{figure}

\begin{figure}[b]
    \begin{center}
    \vspace{-0.2cm}
          \includegraphics[width=0.58\textwidth]{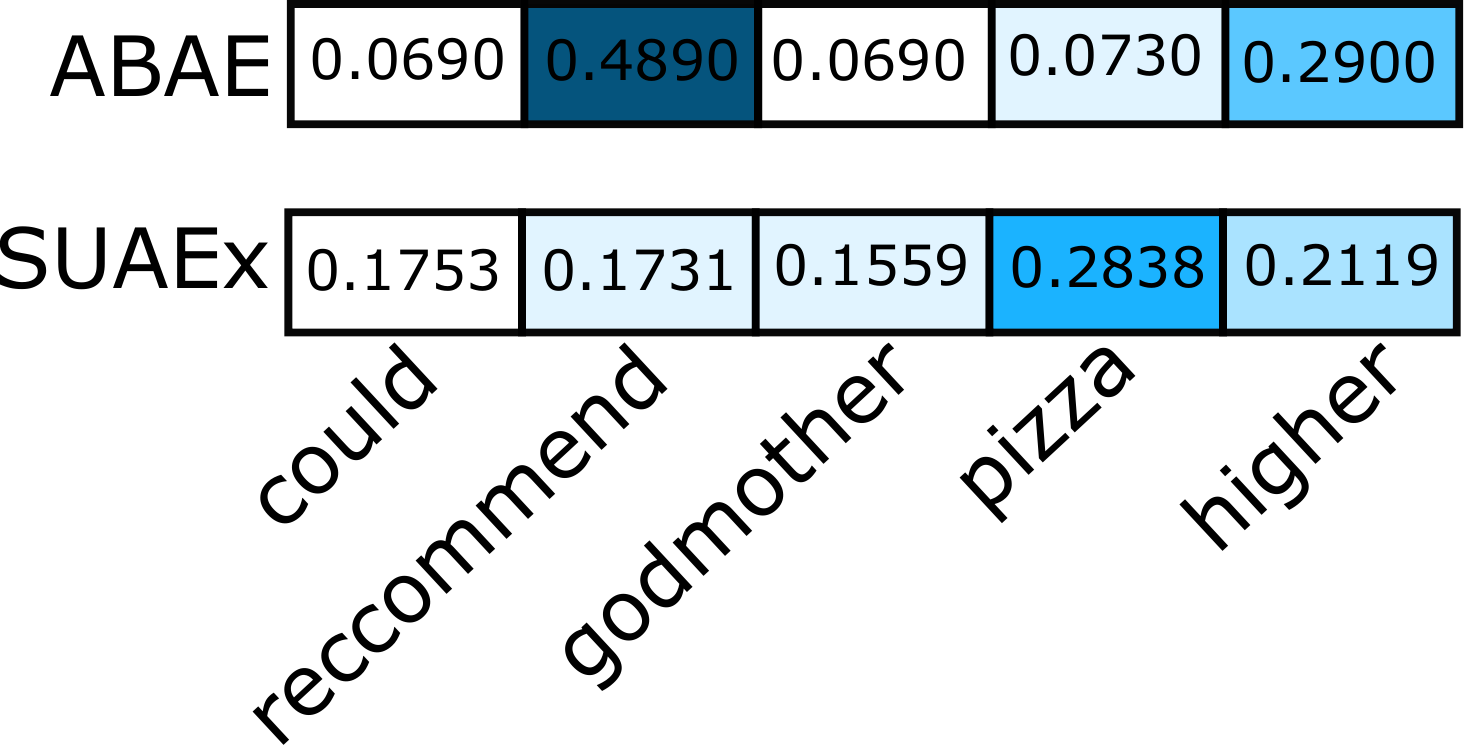}
          \vspace{-0.2cm}
            \caption{An example of the attention values assigned to the input sentence \emph{``I could n't recommend their Godmother pizza any higher"} by the two solutions \label{im:attention_comparison}.}
    \end{center}
        \vspace{-0.2cm}
\end{figure}

\vspace{-0.3cm}
\noindent\textbf{Results for the quality of the aspects and categories extracted}.
The evaluation metrics precision, recall, and F1 were calculated by comparing the outputs generated by both methods against the gold-standard annotations. 
Fig.~\ref{im:overall} shows the results of the evaluation metrics for both approaches averaged across categories.
As an overall tendency, \nome achieves better recall than the baseline in all datasets. ABAE tends to have a better precision in most cases (three out of four datasets). This is expected and can be attributed to the contrast in the way the two solutions use the attention mechanism.
While ABAE only considers the highest attention-valued word in the sentence, \nome uses all the attention values in the sentences. This difference can be seen in the example from Fig.~\ref{im:attention_comparison}. 
 \nome considers the reference words as a type of context to guide category attribution. Basically, for a given sentence, ABAE tries to be precise by focusing on a single word, while \nome tries to be more comprehensive by considering more words.
Our recall improvement was superior to our decrease in precision, so our F1 results were better in all datasets. 
The results demonstrate the adaptability of \nome to different domains. 
On the SemEval datasets, \nome outperformed the baseline in nearly all cases. 
We can attribute ABAE's poor results in the SemEval datasets to the dependence on the training (\ie raw) data. While \nome only uses the training data to generate the word-embeddings representation, ABAE also uses it in the evaluation module because it clusters the training data.

The results of the evaluation metrics per aspect category are shown in Table~\ref{tab:abae_semeval_datasets_result}.  
\nome scored lower in recall for Ambience, Smell, and Look because the reference words (\ie the names of the categories themselves) are not as expressive as the reference words for the other categories. We can find more similar words for general terms like \emph{food} or \emph{staff} than for specific terms like \emph{smell}. 
\begin{table}[b]
\centering
 \caption{\label{tab:abae_semeval_datasets_result} Results for the quality of aspect category extraction.}
 \vspace{-0.2cm}
\begin{footnotesize}
\begin{tabular}{ l | p{1cm} | p{1cm} | p{1cm} || p{1cm} | p{1cm} | p{1cm}}
\toprule
\multicolumn{1}{c|}{\multirow{2}{*}{\textbf{Category}}} & \multicolumn{3}{c||}{\textbf{SUAEx}}         & \multicolumn{3}{c}{\textbf{ABAE}}         \\
\cmidrule(r){2-7}
\multicolumn{1}{c|}{}     & \multicolumn{1}{c|}{\textbf{P}} & \multicolumn{1}{c|}{\textbf{R}} & \multicolumn{1}{c||}{\textbf{F1}} & \multicolumn{1}{c|}{\textbf{P}} & \multicolumn{1}{c|}{\textbf{R}} & \multicolumn{1}{c}{\textbf{F1}} \\
\midrule
\multicolumn{7}{c}{\textbf{CitySearch}}                       \\
\midrule
Food     & 0.917            & \textbf{0.900}   & \textbf{0.908}   & \textbf{0.953}   & 0.741    & 0.828    \\
Staff    & 0.660            & \textbf{0.872}   & 0.752    & \textbf{0.802}   & 0.728    & \textbf{0.757}   \\
Ambience & \textbf{0.884}   & 0.546            & 0.675    & 0.815    & \textbf{0.698}   & \textbf{0.740}   \\
\midrule
\multicolumn{7}{c}{\textbf{BeerAdvocate}} \\
\midrule
Feel       & 0.687    & \textbf{0.832}   & 0.753    & \textbf{0.815}   & 0.824    & \textbf{0.816}   \\
Taste       & \textbf{0.656}   & \textbf{0.794}   & \textbf{0.718}   & 0.637    & 0.358    & 0.456    \\
Smell       & \textbf{0.689}   & 0.614    & \textbf{0.649}   & 0.483    & \textbf{0.744}   & 0.575    \\
Taste+Smell & 0.844    & \textbf{0.922}   & \textbf{0.881}   & \textbf{0.897}   & 0.853    & 0.866    \\
Look        & 0.876    & 0.849    & 0.862    & \textbf{0.969}   & \textbf{0.882}   & \textbf{0.905}   \\
\midrule
\multicolumn{7}{c}{\textbf{Sem2015-Restaurant}} \\
\midrule
Food      & \textbf{0.953}    & \textbf{0.674}    & \textbf{0.789}    &  0.573   & 0.213    & 0.311    \\
Staff     & \textbf{0.882}    & \textbf{0.714}    & \textbf{0.789}    & 0.421    & 0.159    & 0.213    \\
Ambience  & \textbf{0.627}    & \textbf{0.967}    & \textbf{0.760}    & 0.107    & 0.206    & 0.141    \\
\midrule
\multicolumn{7}{c}{\textbf{Sem2015-Laptop}}\\
 \hline
Price    &   0.750   &   \textbf{0.915}   &   \textbf{0.824}   &   \textbf{0.895}   &   0.576   &   0.701  \\
Hardware &   0.785   &   \textbf{0.797}   &   \textbf{0.791}   &   \textbf{0.914}   &   0.481   &   0.631  \\
Software &   \textbf{0.714}   &   \textbf{0.455}   &   \textbf{0.556}   &   \textbf{0.714}   &   0.114   &   0.196  \\
Support  &   0.667   &   \textbf{0.800}   &   \textbf{0.727}   &   0.083   &   0.200   &   0.118  \\
 \bottomrule
\end{tabular}
\end{footnotesize}
\end{table}

Other works have also used the CitySearch and BeerAdvocate datasets. Thus we are also able to compare \nome to them. These works have applied techniques such as LDA~\cite{Zhao2010}, biterm topic-models~\cite{Yan2013}, a statistical model over seed words~\cite{Mukherjee2012}, or restricted Boltzmann machines~\cite{wang2015}. The same tendency found in the comparison with the baseline remains, \ie \nome achieves better recall in all cases except for the categories \emph{Ambience} and \emph{Smell}. In terms of F1, \nome is the winner for \emph{Food}, \emph{Staff}, \emph{Taste}, \emph{Smell}, and \emph{Taste+Smell}.

Fig.~\ref{im:attention_comparison} presents an example in which \nome assigns more accurate attention-values to an input sentence. This happens because the word \emph{pizza} is more similar to our desired category $food$ than the word \emph{recommend}. However, our method is dependent on the reference words, and in some cases it can assign high attention values to adjectives (which typically are not aspects). This happens with the word \emph{higher} in our example, which is an adjective and received the second highest score.
This could be mitigated with a post-filter based on part-of-speech tagging.

In Table~\ref{tab:aspects_bycategory_abae_rest}, we show the aspect words extracted for the CitySearch dataset. The extraction was performed by selecting the highest attention-valued words of each sentence and by considering the category classification results. This extraction can be used as an additional module in our framework (Fig.~\ref{d_suaex}).
\begin{table}[h!]
    \centering
    \caption{\label{tab:aspects_bycategory_abae_rest} Aspects extracted by \nome for the CitySearch dataset}
    \vspace{-0.2cm}
    \begin{footnotesize}
     \begin{tabular}{l|m{10.4cm}}
    \hline
         \bf Category & \bf Aspect Terms \\
        \hline
         \bf Staff & \emph{replied, atmosphere, answering, whoever, welcomed, child, murray, manager, existence, cold, staff, busy, forward, employee, smile, friendly, gave, woman, dessert, early, kid, lady, minute, bar, helpful, wooden, always, greeting, server, notified, busier, nose, night, guy, tray, seating, everyone, hour, crowd, people, seat, lassi, proper, divine, event, folk, even, waitstaff, borderline, ice}\\
        \hline
         \bf Ambience & \emph{antique, atmosphere, outdoor, feel, proximity, scene, sleek, bright, weather, terrace, dining, surroundings, music, calm, peaceful, ambience, cool, sauce, location, ceiling, garden, painted, relaxed, dark, warm, artsy, excellent, tank, furniture, dim, bar, romantic, level, inside, parisian, air, architecture, aesthetic, adorn, beautiful, brightly, neighborhood, ambiance, alley, elegant, decour, leafy, casual, decor, room} \\
        \hline 
         \bf Food & \emph{seasonal, selection, soggy, chinese, cheese, penang, dosa, doughy, corned, sichuan, mojito, executed, innovative, dish, chicken, calamari, thai, butternut, bagel, northern, vietnamese, paris, menu, technique, dumpling, dhal, better, location, congee, moules, rice, sauce, ingredient, good, straightforward, mein, food, dessert, overdone, appetizer, creatively, fusion, know, unique, burnt, minute, panang, risotto, shabu, roti}\\
    \hline
    \end{tabular}
        \end{footnotesize}
\end{table}

\clearpage
\noindent\textbf{Runtime Results}.
In order to obtain the runtime results, we ran both methods on the same configuration (Intel Core i7-4790 CPU, GeForce GTX 745 GPU, and 16GB of main memory). Since both methods went through the same preprocessing steps, our comparison focused on the attention mechanisms (\emph{Attention Neural Network training} in ABAE and the \emph{Similarity as Attention Mechanism} module in \nome).
Table~\ref{tab:runtimes_results} shows the runtime results.
The differences are remarkable -- ranging from one thousand to almost ten thousand times. 
For the BeerAdvocate dataset, we were unable to obtain the runtimes for the baseline because the number of training sentences was too large.
One could argue that, in practice, these differences are not so significant because they concern the training phase, which can be performed once only. However, training has to be repeated for each domain and, from time to time, to cope with how the vocabulary changes.

\begin{table}[h!]
    \centering
    \vspace{-0.3cm}
        \caption{Runtime for both methods in all datasets}
    \begin{tabular}{l|c|c|c|c}
        \hline
        \hline
                        & \bf CitySearch        & \bf BeerAdvocate           & \bf Sem2015-Restaurant       & \bf Sem2015-Laptop            \\ 
        \hline
         \nome          & 42 sec   & 7 min & 13 sec  & 36 sec   \\
         {ABAE}         & 12 hours   & undefined        &  12 hours      & 98 hours \\ \hline
    \end{tabular}
\vspace{-0.3cm}
    \label{tab:runtimes_results}
\end{table}

\vspace{-0.4cm}
\section{Conclusion}
\vspace{-0.3cm}
This paper proposes \nome which is an alternative approach to deep neural network solutions for unsupervised aspect extraction. \nome relies on the similarity of word-embeddings and on reference words to emulate the attention mechanism used by attention neural networks.

With our experimental results, we concluded that \nome achieves results that outperform the state-of-the-art in ABSA in a number of cases at remarkably lower runtimes. In addition, \nome is able to adapt to different domains.

Currently, \nome is limited to dealing with aspects represented as single words. As future work, we will extend it to treat compound aspects such as \emph{``wine list", ``battery life"}. 
Also, we will improve the selection of reference words by using hierarchical data such as subject taxonomies.

\vspace{0.4cm}
\noindent\textbf{Acknowledgments:} 
This work was partially supported by CNPq/Brazil and by CAPES Finance Code 001.
\vspace{-0.2cm}

\bibliographystyle{splncs}
\bibliography{bibs}

\end{document}